\documentclass[10pt]{article} 

\usepackage[preprint]{rlj} 

\usepackage{amssymb}            
\usepackage{mathtools}          
\usepackage{mathrsfs}           
\usepackage{graphicx}           
\usepackage{subcaption}         
\usepackage[space]{grffile}     
\usepackage{url}                
\usepackage{lipsum}             
\usepackage{wrapfig}
\usepackage{dsfont}
\usepackage{booktabs}
\usepackage{natbib}

\newcommand{\squishlist}{
  \begin{list}{\small $\bullet$}
    { \setlength{\itemsep}{0pt}      \setlength{\parsep}{4pt}
      \setlength{\topsep}{1pt}       \setlength{\partopsep}{4pt}
      \setlength{\leftmargin}{1.2em} \setlength{\labelwidth}{1em}
  \setlength{\labelsep}{0.5em} } }
  \newcommand{\squishend}{
\end{list}  }

\newcommand{\xhdr}[1]{\vspace{0pt}\textbf{#1.}}

\def\comment{1}
\if\comment1
\newcommand{\sid}[1]{\textcolor{red}{[Sid: #1]}}
\newcommand{\sn}[1]{\textcolor{blue}{[SN: #1]}}
\newcommand{\cj}[1]{\textcolor{orange}{[CJ: #1]}}
\newcommand{\rp}[1]{\textcolor{purple}{[RJ: #1]}}
\else
\newcommand{\sid}[1]{}
\newcommand{\sn}[1]{}
\newcommand{\cj}[1]{}
\newcommand{\rp}[1]{}
\fi

\title{Improving Human Performance with Value-Aware Interventions: A Case Study in Chess}

\setrunningtitle{Improving Human Performance with Value-Aware Interventions: A Case Study in Chess}

\author{Saumik Narayanan\textsuperscript{1}, Raja Panjwani\textsuperscript{2,3}, Siddhartha Sen\textsuperscript{2}, Chien-Ju Ho\textsuperscript{1}}

\emails{saumik@wustl.edu, rp1223@georgetown.edu, sidsen@microsoft.com, chienju.ho@wustl.edu}

\affiliations{
  $^{1}$\textbf{Washington University in St. Louis}\\
  $^{2}$\textbf{Microsoft Research}\\
  $^{3}$\textbf{Georgetown University}\\
}

\contribution{
We introduce a framework for identifying value-aware intervention opportunities in sequential decision-making tasks using discrepancies between a human policy and its associated value function. In the single-intervention setting, the intervention that maximizes the human value function emerges naturally from the formulation, and we propose a tractable heuristic for settings with multiple interventions under a budget constraint.
}{
\citet{yu2022environment}, studied environment design with stylized human behavior models. \citet{bastani2026improving} provided recommendations using optimal action strategies, analgous to one of our baselines.
}

\contribution{
This paper demonstrates that value-aware intervention strategies can be implemented using data-driven models of human decision-making and evaluates the approach through large-scale simulations and a human-subject study in the domain of chess.
}{
Prior work evaluates interventions primarily in simplified environments. We study a complex sequential decision domain with large-scale human behavioral data, strong AI baselines, and precise notions of human skill-levels, enabling both realistic simulation and human-subject evaluation.
}


\summary{Artificial intelligence systems are increasingly used to assist humans in sequential decision-making tasks. A common approach is to recommend actions that are optimal according to a strong model of the environment. Such recommendations, however, implicitly assume that the decision maker will continue to act optimally afterward. Human decision makers often deviate from optimal play, meaning that actions that are optimal in isolation may lead to states that are difficult for humans to navigate successfully. This raises a fundamental question: how should AI systems intervene when assisting suboptimal decision makers in sequential environments?
In this work, we study value-aware interventions that account for how humans are likely to behave after receiving assistance. Our approach builds on a basic principle from reinforcement learning: when a policy is suboptimal, discrepancies arise between the actions taken by the policy and those that maximize the expected value of future outcomes. We show how such discrepancies can be used to identify opportunities where interventions improve downstream performance. To operationalize this idea, we develop a data-driven framework that learns models of human decision-making from large datasets and uses these models to guide intervention decisions. We study these ideas in the domain of chess, which provides large-scale human gameplay data and strong AI systems for evaluating intervention strategies.}

\begin{document}

\maketitle  

\begin{abstract}
AI systems are increasingly used to assist humans in sequential decision-making tasks, yet determining when and how an AI assistant should intervene remains a fundamental challenge. A potential baseline is to recommend the optimal action according to a strong model. However, such actions assume optimal follow-up actions, which human decision makers may fail to execute, potentially reducing overall performance. In this work, we propose and study value-aware interventions, motivated by a basic principle in reinforcement learning: under the Bellman equation, the optimal policy selects actions that maximize the immediate reward plus the value function. When a decision maker follows a suboptimal policy, this policy–value consistency no longer holds, creating discrepancies between the actions taken by the policy and those that maximize the immediate reward plus the value of the next state. We show that these policy–value inconsistencies naturally identify opportunities for intervention. We formalize this problem in a Markov decision process where an AI assistant may override human actions under an intervention budget. In the single-intervention regime, we show that the optimal strategy is to recommend the action that maximizes the human value function. For settings with multiple interventions, we propose a tractable approximation that prioritizes interventions based on the magnitude of the policy–value discrepancy. We evaluate these ideas in the domain of chess by learning models of human policies and value functions from large-scale gameplay data. In simulation, our approach consistently outperforms interventions based on the strongest chess engine (Stockfish) in a wide range of settings. A within-subject human study with 20 players and 600 games further shows that our interventions significantly improve performance for low- and mid-skill players while matching expert-engine interventions for high-skill players.

\end{abstract}

\section{Introduction}

AI systems are increasingly used to assist humans in complex decision-making tasks, ranging from strategic planning and navigation to domains such as medicine, programming, and finance \citep{esteva2017dermatologist,vaithilingam2022expectation,fischer2018deep}. In many of these settings, decisions unfold sequentially: early choices influence the states encountered later and ultimately determine the final outcome. Designing AI systems that effectively assist humans in such environments raises a fundamental question: when and how should an AI assistant intervene in the decision process?

A natural baseline is to recommend the action that would be optimal according to a strong expert model. For example, in chess, an AI assistant could simply suggest the move preferred by a top engine such as Stockfish. However, optimal actions are typically evaluated under the assumption that the decision maker will continue to act optimally in the future. Human decision makers, however, often deviate from optimal play. As a result, actions that are optimal under perfect continuation may lead to states that are difficult for humans to navigate, increasing the likelihood of future mistakes. Conversely, actions that appear objectively suboptimal may lead to trajectories that are easier for humans to execute successfully, ultimately yielding better overall outcomes.

This intuition can be formalized through a basic principle in reinforcement learning. Under the Bellman equation, the optimal policy selects the action that maximizes the immediate reward plus the value of the next state. In other words, the policy and the value function satisfy a consistency condition: the action chosen by the policy maximizes the expected value of future outcomes \citep{puterman1994markov}. When a decision maker follows a suboptimal policy, this policy–value consistency no longer holds. Rather than viewing this discrepancy purely as a theoretical artifact, we use it as a signal for identifying opportunities where interventions can improve downstream performance.

Building on this idea, we study value-aware interventions for assisting suboptimal decision makers in sequential environments. Instead of recommending the action that maximizes the value under optimal play, the assistant recommends the action that maximizes the expected outcome when the trajectory continues under the human’s policy. We formalize this problem as a Markov decision process in which an AI assistant may override human actions under a limited intervention budget. This formulation captures a common practical constraint: interventions may be costly due to cognitive load, reduced autonomy, or system design considerations, making selective intervention important.

A key challenge in applying this approach is that the human policy and its associated value function are generally unknown. In this work, we address this challenge using a data-driven approach. Leveraging large datasets of human decision trajectories, we learn models that approximate both how humans act and how their actions translate into expected outcomes. These learned models allow us to estimate policy–value discrepancies and identify states where interventions are most likely to improve performance.

We study this framework in the domain of chess, which provides a particularly useful testbed for human-AI decision support, with several key advantages: First, large-scale datasets of human gameplay are available, enabling accurate modeling of human policies across a wide range of skill levels \citep{lichessdb}. Second, chess has extremely strong AI systems that can approximate optimal play, providing reliable baselines for comparison \citep{stockfish}. Third, recent work has demonstrated that neural networks can learn models that closely mimic human decision-making patterns, enabling realistic simulations of how humans are likely to act in future positions \citep{mcilroy2020aligning}. Together, these properties make chess an ideal environment for studying how AI assistance should adapt to human behavior in sequential tasks.

Using large-scale datasets of online chess games, we train behavioral cloning models that estimate both human policies and their associated value functions across different skill levels. With these models, we analyze intervention strategies under both single- and multiple-intervention settings. Our results show that interventions designed to maximize the expected outcome under human play can substantially improve performance when interventions are limited. In particular, we find that these interventions outperform recommendations based purely on the strongest chess engine for at all skills when the number of interventions is small. We further validate these findings through a human-subject experiment involving 20 players and 600 games, showing that interventions significantly improve performance for low-skill and medium-skill players while matching expert interventions for high-skill players.

Overall, this work makes the following main contributions.
First, we introduce a framework for identifying intervention opportunities in sequential decision-making tasks using discrepancies between human policies and value functions.
Second, we derive value-aware intervention strategies under both single-intervention and multiple-intervention settings.
Third, we demonstrate that such strategies can be implemented in practice using data-driven models of human behavior and validate their effectiveness through large-scale simulations and a human-subject study in the domain of chess.

\section{Related Work}

\xhdr{Human Interventions in Sequential Environments} Our work most closely relates to a line of work on sequential interventions on human decision making, which looks at how to best intervene on a human's trajectory of decisions in order to improve their performance. \citet{yu2022environment} formulate this as an environment design problem and propose to either update the environment rewards or make ``action nudges'' to change human behavior under an intervention budget, and show that human-aware interventions improve performance in a gridworld environment under theoretical models of human behavior. Our approach directly builds on this work, by showing how to apply this framework in real-world domains using data-driven models of human behavior.

A different approach is taken by \citet{bastani2026improving}, who show that, given human trajectories in an Overcooked environment, selecting the intervention with the highest expected difference between the human and an optimal agent leads to improved performance. However, their approach does not account for the impact of suboptimal human policies on the estimation of value and Q functions. One of the main contributions of our approach, in contrast, is the use of human value-aware interventions that explicitly account for how humans are expected to perform after the intervention, rather than assuming optimal human behavior. Moreover, their method relies on the (limited) observed trajectories to empirically estimate Q functions, whereas we leverage data-driven methods to estimate both the policy and value functions.

Alternative notions of interventions can be seen in \citet{nofshin2024reinforcement}, who modify the behavioral clone's MDP parameters (e.g. discount factor, costs, rewards), rather than their actions, and in \citet{straitouri2025narrowing}, who reduce the action space rather than directly selecting actions. Another line of work looks at the effect of AI recommendations (rather than direct interventions) on human decision making, including in sequential decision making settings. For instance, \citet{chen2023learning} and \citet{grand2026best} both analyze settings where AI recommendations are not fully followed, and learn to provide recommendations which are are robust to partial adherence.

\xhdr{Modeling Human Behavior}
There are a variety of approaches to modeling human behavior in sequential decision making settings. We use behavioral cloning \citep{bain1995framework}, a method to learn a supervised model from data which is widely used in domains such as robotics \citep{florence2022implicit} and autonomous driving \citep{wang2022high}. Inverse Reinforcement Learning is an alternative approach, which learns the underlying reward function of human behavior, rather than learning the policy directly \citep{ng2000algorithms}.

Specifically in chess, there have been a number of recent works which have looked at modeling human behavior. \citet{mcilroy2020aligning} show that a behavior cloning approach over a large human dataset can successfully predict actions, and this has been followed up with works showing that individualized models can even better predict behavior when accounting for playing style \citep{mcilroy2022learning}. Other works have looked at adding search to further improve modeling performance \citep{jacob2022modeling,zhang2024human}, and \citet{narayanan2022improving} studied approaches to learning strong models with very small human data budgets.

\section{Setting and Methodology}
\label{sec:formulation}

\subsection{Problem Setting}

Our work is situated in the context of a Markov decision process (MDP), defined by the tuple 
$(\mathcal{S}, \mathcal{A}, \mathcal{P}, \mathcal{R}, \gamma, \mathcal{S}_0)$, where $\mathcal{S}$ is the state space, $\mathcal{A}(s)$ is the action space, $\mathcal{P}(s,a)$ is the transition function, $\mathcal{R}(s,a)$ is the reward function, $\gamma$ is the discount factor, and $\mathcal{S}_0$ is the distribution of starting states. In this paper, we focus on finite MDPs with $\gamma = 1$ and episode length $T$, though our discussion generalizes to other MDPs as well.

A (stochastic) policy is denoted by $\pi(a|s)$. Let $\zeta = \{(s_t,a_t,r_t)\}_{t=1}^T$ denote a trajectory generated by policy $\pi$ starting from an initial state $s_0$. The value function of a policy is defined as
$
V^\pi(s) = \mathbb{E}_{\zeta \sim \mathcal{Z}^\pi(s)}\big[\sum_{t=1}^T r_t\big],
$
where $\mathcal{Z}^\pi(s)$ denotes the distribution over trajectories induced by policy $\pi$ starting from state $s$. The expected reward of a policy is
$
J(\pi) = \mathbb{E}_{s_0 \sim \mathcal{S}_0}\big[V^\pi(s_0)\big].
$
The Q-function of a policy is defined as
$
Q^\pi(s,a) = \mathbb{E}_{s' \sim \mathcal{P}(\cdot|s,a)}\big[\mathcal{R}(s,a) + V^\pi(s')\big].
$

In this work, our goal is to design an intervention policy that improves the expected performance of the human decision maker while limiting the frequency of interventions. 
We assume access to a human policy $\pi_H$, and our task is to learn an \emph{intervention policy} consisting of (i) a \emph{gating function} $\phi(s)\in[0,1]$ that denotes the probability of overriding the human at state $s$, and (ii) an \emph{override policy} $\pi_I(a|s)$ that specifies the action taken when an intervention occurs. 
We use $\pi_H \oplus (\phi,\pi_I)$ to denote the policy after intervening human policy $\pi_H$ with intervention $(\phi,\pi_I)$: 
\begin{equation}
(\pi_H \oplus (\phi,\pi_I))(a|s) =
\begin{cases}
\pi_I(a|s) & \text{with probability } \phi(s) \\
\pi_H(a|s) & \text{with probability } 1-\phi(s).
\end{cases}
\label{eq:post-intervention}
\end{equation}

Under the constraint that the expected intervention rate lies under some budget frequency $B \in [0,1]$, denoting the maximum expected fraction of timesteps at which the AI may override the human policy, our goal is to find the optimal intervention policy and gating function:
\begin{align}
\arg\max_{\phi, \pi_I} J(\pi_H \oplus (\phi,\pi_I))
\quad
\text{s.t.}
\quad
\frac{1}{T}\,\mathbb{E}_{\zeta\sim\mathcal{Z}^{\pi_H \oplus (\phi,\pi_I)}}
\left[\sum_{t=1}^T \phi(s_t)\right] \le B .
\label{eq:intervention-design}
\end{align}

\subsubsection{A Case Study in Chess}

In this work we focus on the domain of chess as a case
study. Chess provides a rich sequential environment with well-defined rules,
large-scale datasets of human play, and strong AI systems that allow us to
analyze intervention strategies in a realistic setting. In chess, the elements of the MDP correspond naturally to the game
structure. A state $s \in \mathcal{S}$ corresponds to a board position, an action
$a \in \mathcal{A}(s)$ corresponds to a legal move available in that position,
and the transition function $\mathcal{P}(s,a)$ is determined by the rules of
chess. 
The reward function reflects the final game outcome, typically defined
as $1$ for a win, $0.5$ for a draw, and $0$ for a loss from the perspective of
the player.

A key component of our framework is the human policy $\pi_H(a | s)$ and its
associated value function $V^{\pi_H}(s)$, which capture how a human player
behaves and the expected outcome when play continues according to that behavior.
In practice, we do not have direct access to these quantities. Instead, we learn
an approximation using behavioral cloning trained on large datasets of human
games, using the methodology by \citet{mcilroy2020aligning}. The behavioral cloning model provides (i) a policy head that predicts the
distribution of human moves $\pi_H(a|s)$ and (ii) a value head that
estimates the expected outcome when the trajectory continues under the learned
human policy, providing an estimate of $V^{\pi_H}(s)$.  For comparison with optimal play, we use the chess engine Stockfish to approximate the optimal policy $\pi^*$ and the associated optimal value function $V^{\pi^*}(s)$.

\subsection{Value-Aware Intervention Design}
We now describe our approach for designing value-aware interventions. The key idea is that when assisting a human decision maker in a sequential decision-making environment, the assistant should not necessarily recommend the action from the optimal policy. Actions that are optimal under perfect play may lead to states that are difficult for humans to navigate, resulting in mistakes later in the trajectory. Instead, the assistant should recommend the action that maximizes the expected downstream performance of the human who will continue the trajectory. As described in Section~3.1.1, we approximate the human policy $\pi_H(a | s)$ and the associated value function $V^{\pi_H}(s)$ using a behavioral cloning model trained on human gameplay data.

In reinforcement learning, optimal policies satisfy the Bellman optimality condition, which states that the optimal action maximizes the expected immediate reward plus the value of the next state. As a result, the optimal policy $\pi^*(a|s)$ selects actions that maximize $Q^{\pi^*}(s,a)$, ensuring consistency between the policy and the value function. Human decision makers, however, often follow suboptimal policies, and therefore there can be discrepancies between the actions by the human policy and those that would maximize the expected value of future outcomes.
We use these discrepancies to identify opportunities for intervention. Rather than selecting actions that lead to the objectively highest-value states (as done by strong chess engines such as Stockfish), interventions should be chosen based on how they affect the expected outcome when the human continues the trajectory. Formally, this corresponds to evaluating candidate actions using the human value function $V^{\pi_H}$ and the associated action-value function $Q^{\pi_H}$.

\subsubsection{Single-Intervention Regime}
First, we consider the setting where at most one intervention is allowed per trajectory. In this setting, there is a simple solution for the optimal intervention when we have access to $\pi_H$, and $V^{\pi_H}$. \footnote{If we have access to value function $V^{\pi_H}$, we can derive $Q^{\pi_H}$ by adding the immediate reward.}

When at most one intervention is allowed per episode, the post-intervention behavior follows $\pi_H$ by definition. Consequently, the value of taking action $a$ at state $s$ is given by $Q^{\pi_H}(s,a)$, since the remainder of the trajectory follows the human policy.
In this regime, the benefit of overriding the human at state $s$ with action $a$ 
is
$
  \Delta^{\pi_H}(s,a) = Q^{\pi_H}(s,a) - V^{\pi_H}(s),
$
which measures the improvement in expected outcome relative to allowing the human to act.
The optimal override action at state $s$ for policy $\pi_H$ is therefore
$  
a(\pi_H, s) = \arg\max_a Q^{\pi_H}(s,a).
$
Similarly, over the course of a trajectory $\zeta$ generated by $\pi_H$, the optimal state to intervene is the state that maximizes this improvement:
  $s(\pi_H, \zeta) = \arg\max_s \big[ \max_a \Delta^{\pi_H}(s,a) \big].$

Thus, in the single-intervention setting, $Q^{\pi_H}$ determines both (i) the best action to recommend and (ii) the optimal criterion for selecting the intervention state. Given a full human trajectory $\zeta$, the optimal intervention $(\phi^*,\pi_I^*)$ that solves Equation~\ref{eq:intervention-design} satisfies that (1) $\phi^*(s)=1$ if $s=s(\pi_H,\zeta)$ and $\phi^*(s)=0$ otherwise, and (2) $\pi^*_I(a|s)=1$ if $a=a(\pi_H,s)$ and $\pi^*_I(a|s)=0$ otherwise.

In practice, the realized trajectory might not be known in advance, and the assistant must decide whether to intervene as the trajectory unfolds. In this case, the problem reduces to a standard optimal stopping problem \citep{peskir2006optimal}. 

\subsubsection{Multiple-Intervention Regime}
\label{sec:formulation-multi}

The above intervention methodology is optimal when at most one intervention is allowed per trajectory. When multiple interventions are permitted, however, selecting actions based on $Q^{\pi_H}(s,a)$ is no longer guaranteed to be optimal. The reason is that $Q^{\pi_H}$ assumes the remainder of the trajectory follows the human policy $\pi_H$, whereas in a multiple-intervention setting additional interventions may occur later, causing the resulting policy to deviate from $\pi_H$.

Ideally, the value of an action should therefore be evaluated under the post-intervention policy $\bar{\pi} = \pi_H \oplus (\phi, \pi_I)$. This leads to a significantly more challenging optimization problem than in the single-intervention setting, since the value function $Q^{\bar{\pi}}$ depends on the intervention policy $(\phi,\pi_I)$ itself.
To obtain a tractable procedure, we approximate $Q^{\bar{\pi}}$ using $Q^{\pi_H}$. This approximation is reasonable when the intervention budget $B$ is relatively small. When interventions occur infrequently, the resulting trajectory distribution remains close to the human trajectory distribution induced by $\pi_H$. Under this approximation, we evaluate intervention opportunities using the same improvement signal as in the single-intervention setting, $\Delta^{\pi_H}(s,a) = Q^{\pi_H}(s,a) - V^{\pi_H}(s).$
This yields the simplified optimization problem. The override policy $\pi_I^*$ follows the same solution as in the single-intervention regime:  $\pi^*_I(a|s)=1$ if $a=a(\pi_H,s)$ and $\pi^*_I(a|s)=0$ otherwise. 
The gating function $\phi^*$ is obtained by solving
\begin{equation*}
\begin{aligned}
\arg\max_{\phi}\;
\mathbb{E}_{s \sim \zeta^{\pi_H}}
\left[
\phi(s)\max_a \Delta^{\pi_H}(s,a)
\right]
\quad
\text{s.t.}\;
\mathbb{E}_{s \sim \zeta^{\pi_H}}[\phi(s)] \le B .
\end{aligned}
\end{equation*}

Although this approach relies on approximating $Q^{\bar{\pi}}$ with $Q^{\pi_H}$, which is most accurate when $B$ is small, we find empirically that it performs well even for larger intervention budgets. In our experiments, the method continues to outperform baseline policies even when we allowing interventions up to 50\% of the time.

\section{Simulation Results}
\label{sec:simulation-results}

In the previous section, we introduced a framework for value-aware interventions and derived intervention strategies for both the single-intervention and multiple-intervention regimes. In particular, our analysis shows that when only a single intervention is allowed, selecting the action that maximizes the human value function $Q^{\pi_H}$ naturally emerges as the optimal intervention. For settings with multiple interventions, we proposed a practical approximation that prioritizes interventions based on the discrepancy between the human policy and its associated value function.

In this section, we empirically evaluate these predictions through a large-scale simulation study in the domain of chess. Chess provides a rich sequential environment with strong AI baselines and large datasets of human gameplay, allowing us to model human behavior and evaluate intervention strategies. In particular, we compare interventions that optimize the expected outcome under the human policy with interventions that select actions based on a strong expert model (Stockfish), which assumes optimal continuation. Using models of human behavior trained on large datasets of online games, we simulate both human play and AI interventions to measure how different strategies affect downstream performance. We evaluate these strategies in both the single-intervention setting and multiple-intervention regimes under limited intervention budgets.

\subsection{Experiment Setup}

For all experiments in this paper, we use a Behavioral Cloning (BC) approach to model human behavior. First, we sample a dataset of 256 million chess positions, actions, and game outcomes from the Lichess dataset \citep{lichessdb}, such that the blitz rating of the players range uniformly between 400 and 2800. We then fine-tune an existing pretrained Leela T82 model on this dataset. Notably, we train a single BC model across all ratings, but include the players' ratings as an input to the model, allowing it to learn a single model which can generalize across all ratings. As a result, we are able to use this BC model to predict both the human policy function $\pi_H$ and the human value function $V^{\pi_H}$ at any given rating. \footnote{We provide more details on model training in Appendix \ref{sec:training-details}, and the full code to reproduce our results is available at https://anonymous.4open.science/r/value-aware-interventions-0C08/.}

\subsection{Single-Intervention Experiments}


To compute the efficacy of single interventions, we leverage the Lichess dataset of human games and simulate interventions at random positions sampled from the dataset. We then evaluated the counterfactual outcome after taking an intervention these positions, and then, we simulate 64 game rollouts from the resulting position using the trained BC model as our approximation of the human policy, setting the policy rating for both sides as the player rating at the initial position.

Then, we evaluated three different classes of intervention strategies in each position:

\textbf{Human Move:} This is the actual move that the human played in the game, and serves as a baseline for comparison.

\textbf{Stockfish Intervention:} We intervene with the move which maximizes the expected value according to Stockfish (e.g. the optimal move assuming optimal follow-up).

\textbf{Valuemax Intervention:} We intervene with the move which maximizes the expected value prediction according to our trained BC model at the human's rating (e.g. the optimal move, assuming human follow-up).

\begin{wrapfigure}{r}{0.48\textwidth}
  \vspace{-10pt}
  \centering
  \includegraphics[width=0.48\textwidth]{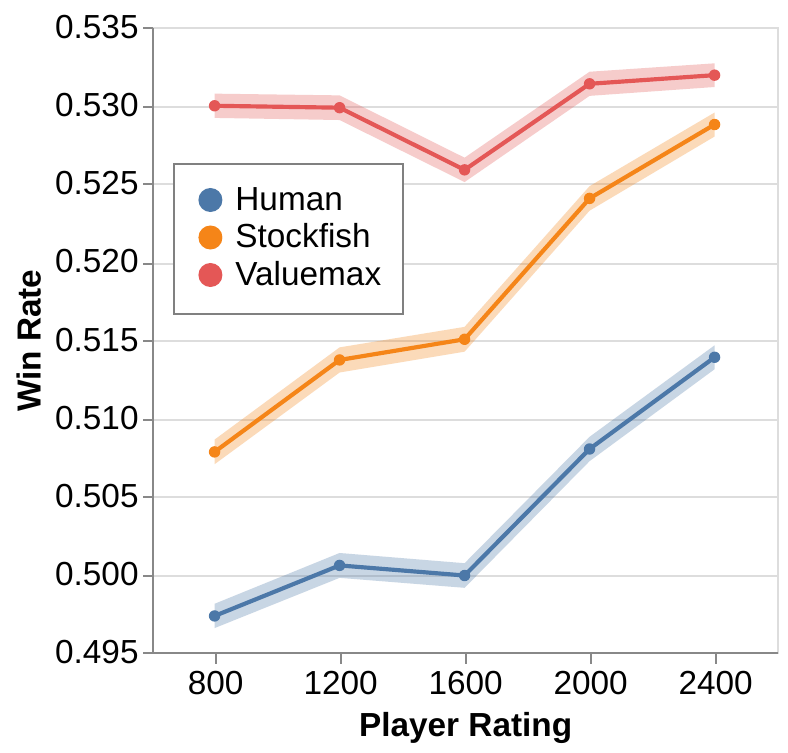}
  \caption{Win rate and standard error of each single-intervention strategy across ratings. }
  \label{fig:single-intervention}
  \vspace{-10pt}
\end{wrapfigure}

We evaluate each of these strategies at 100k random positions from sampled from games played by players at 800, 1200, 1600, 2000, and 2400 ratings (500k positions in total), and report the resulting win rate and standard error of each intervention strategy in Figure~\ref{fig:single-intervention}.

Overall, these results provide empirical validation for our theoretical predictions from Section~\ref{sec:formulation} that Valuemax is the optimal strategy in the single-intervention regime. At all ratings, the Valuemax strategy outperforms both the baseline human move and the Stockfish strategy ($p\!<\!0.001)$. Interestingly enough, the gap between Stockfish and Valuemax tends to strongly decrease as player strength goes up -- for 800-rated players, Valuemax performs over 2\% better than Stockfish, while for 2400-rated players, Valuemax only helps by around 0.3\%. This trend aligns with intuitions, since Valuemax is explicitly exploiting the non-optimality of the human in order to make recommendations. As the human policy tends closer towards perfectly optimal, there is a smaller difference to exploit, so the difference between Valuemax and Stockfish will correspondingly reduce.

\subsection{Multiple-Intervention Experiments}

Next, we consider an intervention regime where we can intervene multiple times during the course of a game. This setting is more practical for real-world deployment, as we may want to provide ongoing assistance to players throughout a game, rather than just on a single move. While our theoretical framework does not provide us with optimality guarantees in this regime, we can still test the extent to which our framework described in Section \ref{sec:formulation-multi} provides empirical improvements in performance over baselines.

In particular, we consider a threshold-based intervention strategy, where we intervene whenever the expected improvement in performance exceeds a certain threshold. This allows us to control the frequency of interventions, using higher thresholds to ensure fewer interventions and lower thresholds for more frequent interventions. At the extreme ends of the spectrum, a threshold of 1 corresponds to a 0\% intervention frequency, while a threshold of 0 corresponds to a 100\% intervention frequency.

Then, we evaluate two intervention strategies, Valuemax and Stockfish across a range of thresholds. Intuitively, we would expect Valuemax to outperform Stockfish when interventions are infrequent, as Valuemax is designed to maximize performance of a policy which acts similarly to an actual human at that skill level. However, as the rate of interventions increases, the more the future policy diverges from the human policy, and the less well we would expect the Valuemax strategy to be calibrated to the resulting mixed policy.

\begin{wrapfigure}{r}{0.44\textwidth}
  \vspace{-10pt}
  \includegraphics[width=\linewidth]{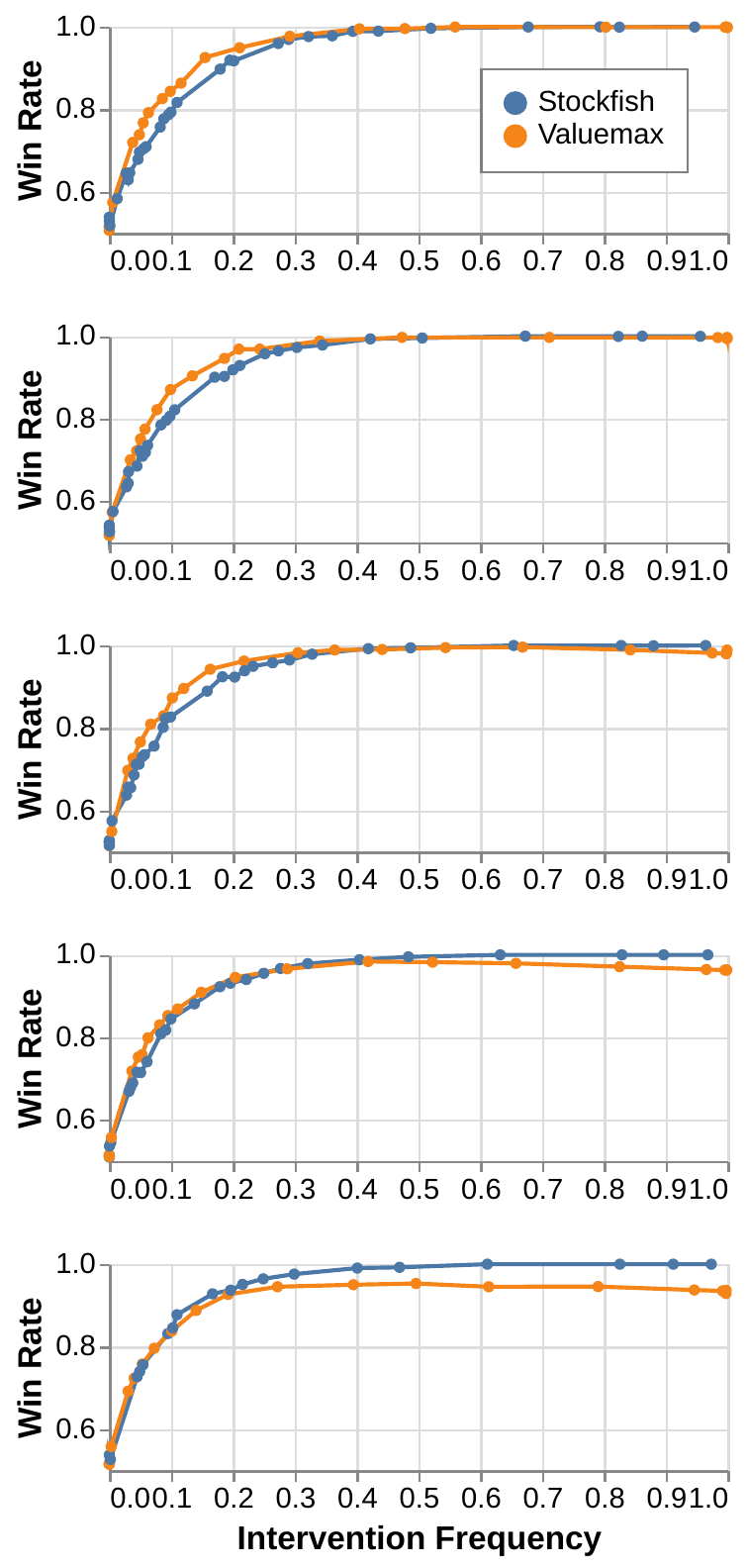}
  \caption{Win rate of each multiple-intervention strategy across player ratings and intervention frequency budgets.}
  \label{fig:online}
  \vspace{-10pt}
\end{wrapfigure}

To evaluate each intervention strategy, we simulate full games starting from the initial chess position. At each move, we compute the improvement over the expected human policy (estimated using our learned BC model $\pi_H$). We then intervene if the expected improvement exceeds the specified threshold and otherwise play the move sampled from the BC model, continuing this process until the game ends. After the game, we record the overall outcome and the total number of interventions made, as a percentage of intervention opportunities. We play 2560 randomized games for each strategy, threshold, and player rating combination, and report the average intervention frequency and win rate in Figure~\ref{fig:online}.



Overall, our results align with our expectations on the relationship between intervention frequency and performance for both strategies. At each player rating, Valuemax outperforms Stockfish at small intervention budgets. For instance, for 800-rated players, Valuemax outperforms Stockfish with budgets as high as $B=0.5$, while for 2400-rated players, this is true for budgets up to $B=0.05$. Conversely, Stockfish outperforms Valuemax for players at each rating with large intervention budgets, as the divergence from the human policy reduces the calibration of Valuemax.

\subsection{Exploring Human-Understandable Concepts}

One advantage of studying interventions in the domain of chess is that positions can be evaluated using a large number of structured, human-understandable concepts derived from classical chess evaluation functions. Modern engines such as Stockfish compute dozens of intermediate features that capture properties of a position such as material balance, piece activity, pawn structure, and king safety. While our intervention strategy itself does not directly use these concepts, they provide a useful lens for examining the types of positions in which interventions occur.

We conduct a limited analysis exploring how these concepts correlate with our proposed intervention strategy, and present the results in Appendix \ref{sec:concepts}. Here, we highlight one preliminary finding -- a weak enemy king correlates very strongly with low-skill interventions, but not with high-skill interventions. One possible explanation of this is that low-skill players need help to checkmate the king while on the attack, while high-skill players already know how to deliver mate, so interventions may not be needed in those positions.

This type of analysis combining value-aware interventions and human-understandable concepts has strong potential to lead to more explainable AI assistants or even insights on human pedagogy (finding positions and concepts where a particular human is weakest). We leave this avenue of analysis to future researchers.

\section{Human Study}

In Section \ref{sec:formulation}, we provided theoretical justifications for the Valuemax intervention strategy, and in Section \ref{sec:simulation-results}, we validated through simulation our approach in both single-intervention and multi-intervention regime. However, our simulated results still rely on evaluations using a synthetic model, which leaves open the question of whether these results hold for real human players.

In this section, we aim to demonstrate using a within-subjects human experiment\footnote{Our experiment design was approved by our institution's IRB.} that interventions using the Valuemax strategy indeed outperform both human policy baselines and interventions which assume that humans will follow-up with optimal behavior.

\subsection{Human Participant Recruitment}


\begin{wraptable}
  {R}{0.26\textwidth}

  \vspace{-40pt}
  \centering
  \begin{tabular}{|r|r|}
    \hline
    \textbf{
      \begin{tabular}[c]{@{}r@{}}Standardized\\Rating
    \end{tabular}} & \textbf{n}  \\
    \hline
    800--1600 & 7 \\
    \hline
    1600--2000 & 6 \\
    \hline
    2000--2400 & 7 \\
    \hline
    \textbf{Total} & \textbf{20} \\
    \hline
  \end{tabular}
  \caption{Summary rating statistics of human participants in our study.}
  \label{tab:human_subject}
\end{wraptable}

By advertising at local chess clubs, we recruited 20 players to take part in our experiment. In order to ensure that participants were chess players, and to give them opponents at roughly the same skill level, we required each participant to have played at least 50 lifetime games on the Lichess or Chess.com online platforms in the Bullet, Blitz, or Rapid time controls. Because each website and each time control has a different rating scale and distribution, rating standardization was required to fairly compare players. To do this, we used the cross-platform rating dataset collected by \citet{jensen2025rating}. In Table~\ref{tab:human_subject}, we summarize the standardized ratings of the players in the study. Each participant spent approximately 60 minutes to complete our study, and we compensated each player with a \$30 Amazon gift card upon completion.

\subsection{Study Design}

In this study, we evaluate three different single-intervention approaches: Human, Stockfish, and Valuemax.
While it would be ideal to evaluate the performance of each intervention strategy across the true distribution of states encountered by players at each rating, this would require an infeasible number of games to be played by human participants, due to the small expected differences in performance between the intervention strategies. For example, in our simulated single-intervention results in Section \ref{sec:simulation-results}, we see that the expected improvement of Valuemax over Stockfish interventions hovers between 0.3\% and 2.5\% across ratings, meaning that we could need tens of thousands of games to have enough statistical power to detect significant differences between the strategies.

Instead, we take a more targeted approach to evaluate the performance of each intervention strategy in states where we would expect to see significant differences in performance between the strategies. In particular, we use our simulated results to identify states where Valuemax interventions are expected to yield significantly higher improvements over Stockfish moves and human moves, and then evaluate the real-world performance of each intervention strategy in these states with human participants. This allows us to test whether the expected improvements predicted by our simulated results translate to real-world improvements when intervening with human players. In particular, we randomly selected 20 positions at each rating where the expected improvement of Valuemax over Stockfish and the BC policy\footnote{Comparing Valuemax to the policy is a more fair comparison here than comparing to the Human move, since in many setups, we may not have access to the human move before we must decide to give a recommendation.} were each at least 20\%, giving us 100 total positions.

Each participant in the user study was randomly assigned 30 positions to evaluate based on their rating, taken from position sets at the two closest ratings to them. For example, if a user had a standardized rating of 1900, we combined the 20 positions at the 1600 level and the 20 positions at the 2000 level, into a pool of 40 positions and then randomly sampled 30 positions to provide to the user.\footnote{This additional buffer allowed us to fill the full quota even if a few games were cut short due to user disconnections.} Then, for each position, we randomly assigned one of the three treatments (Human, Stockfish, Valuemax), such that each participant ended up playing 10 games under each treatment

In this work, we are primarily focused on understanding which strategy leads to the best outcome, assuming that recommendations were taken consistently.
Therefore, in our study design, we presented players with the board position \emph{following} the recommendation move, having the user continue the rest of the game under a standard blitz time control of 3+2, playing against the BC policy model at the position rating.

\subsection{Results}

\begin{wrapfigure}{r}{0.48\textwidth}
  \vspace{-50pt}
  \includegraphics[width=0.48\textwidth]{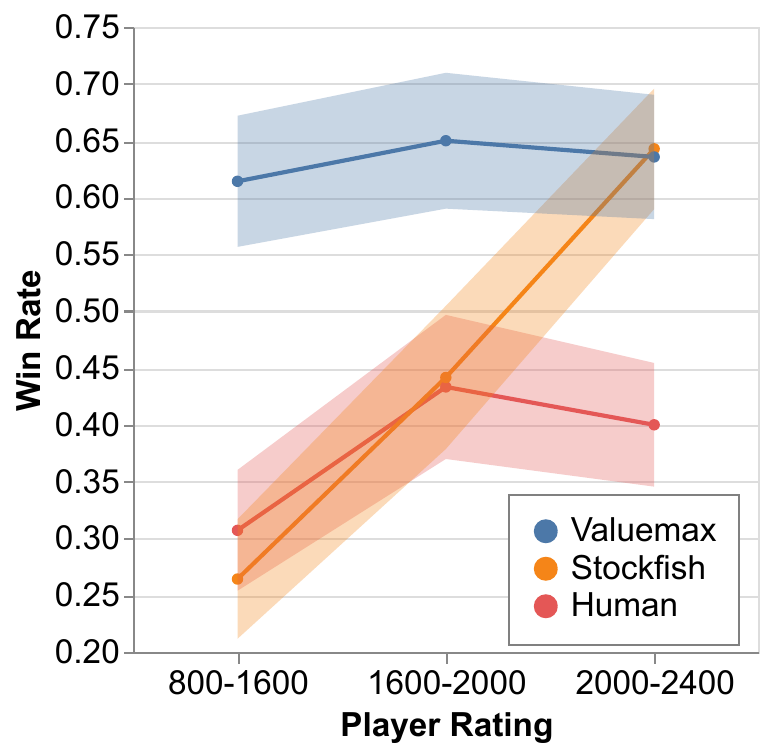}
  \caption{Human performance of each intervention strategy across ratings.}
  \label{fig:human_subject}
  \vspace{-10pt}
\end{wrapfigure}

Overall, we collected data on 600 total games played by 20 human participants across the three different intervention strategies, using a custom chess platform designed for this experiment. We report the win rate for each intervention strategy across player ratings in Figure~\ref{fig:human_subject}.

We find strong evidence that a Valuemax intervention strategy can outperform Stockfish-based interventions and the human baseline, especially at lower ratings. While we cannot directly compare the results to those in Figure \ref{fig:single-intervention} due to the difference in initial state distributions, the results still align with our expectations. Using a Mixed-Effects Linear Model statistical test, Valuemax significantly outperforms both Stockfish and the human baseline at all ratings under 2000 ($p\!<\!0 .001$).
Interestingly, we see no effect between Valuemax and Stockfish at the highest rating levels, despite our initial expectations estimating over a 20\% performance gap between Valuemax and Stockfish at every rating level. On the other hand, at the lowest skill level, we see differences greater than 35\% at the lowest skill level, despite still predicting the same 20\% expected performance gap. Overall, the sharp slope upwards on the Stockfish line indicates that our simulated results might underestimate the Stockfish intervention strategy at high levels, but possibly overestimates the Stockfish intervention strategy at low skill levels.

\section{Conclusion and Discussion}

In this work, we study how AI systems should intervene when assisting humans in sequential decision-making tasks. While a natural baseline is to recommend the action from a strong expert model, such recommendations implicitly assume optimal continuation. When human decision makers deviate from optimal play, these recommendations may lead to states that are difficult for them to navigate, reducing overall performance. We propose value-aware interventions that instead select actions based on the expected outcome when the trajectory continues under the human policy. Our analysis shows that when at most one intervention is allowed, recommending the action that maximizes the human value function naturally arises as the optimal intervention. For settings with multiple interventions, we propose a practical approximation that prioritizes interventions based on discrepancies between the human policy and value functions.

We evaluate these ideas in the domain of chess using data-driven models of human behavior trained on large datasets of gameplay. In both simulation and a human study, interventions that account for human behavior outperform recommendations based solely on optimal play when interventions are limited, particularly for low- and mid-skill players. These results support the intuition that actions optimal under perfect play may not be optimal when the decision maker follows a different policy. We also observe that the advantage of human-aware interventions depends on the intervention budget: when interventions are rare, accounting for human behavior yields clear improvements, whereas as the intervention frequency increases the resulting trajectories diverge from the human policy, making optimal-play recommendations increasingly competitive. Finally, we conduct a preliminary analysis exploring correlations between value-aware interventions and human-understandable concepts, with potential connections to explainable AI assistance and human pedagogy.

\textbf{Limitations and Future Work} Our work has several limitations. First, the approach relies on learned behavioral models of human decision making, which may not perfectly capture real behavior, especially in domains with more limited data. Second, our simulation experiments depend on rollouts from the behavioral cloning model rather than real human trajectories. Third, our human-subject study assumes that recommendations are followed exactly, whereas in practice users may ignore or partially follow AI suggestions. Finally, our evaluation focuses on chess, a domain with abundant data and strong AI baselines. Future work could explore extending these ideas to other sequential decision-making settings, including domains where human behavior is more heterogeneous and where interventions may interact with learning and skill development over time.

Overall, our findings suggest that AI assistants in sequential environments should account for how users are likely to act after receiving assistance, rather than relying solely on optimal-action recommendations. Incorporating models of human behavior into the design of intervention strategies may therefore play an important role in building AI systems that more effectively support human decision making.




\bibliography{cite}
\bibliographystyle{rlj}

\appendix

\section{Model Training Details}
\label{sec:training-details}

In this section, we provide more details on the BC model used for all simulated analysis in the paper. All code is available at https://anonymous.4open.science/r/value-aware-interventions-0C08/.

\subsection{Dataset Construction}

All data is taken from the Lichess online database, which contains over 7.5 billion chess games played by humans. We sample from blitz games played between January 2024 and November 2025. We construct a dataset of positions taken from games played by players rated between 400 and 499. We then construct similar datasets at 500-599, etc., all the way to 2800-2899. We then uniformly sample from each of these datasets to construct our final, mixed-skill training dataset.

For all validation, testing, and base positions for our simulated results, we follow a similar dataset construction process with games played in December 2026.

\subsection{Model Training}

Our model is initialized with the publicly available, pre-trained Leela T82 architecture \citep{lc0} with 768 channels, 15 blocks, and 24 attention heads. The policy head outputs an 1858-dimensional move distribution, and the value head predicts win/draw/loss probabilities.

We then train the model on the constructed dataset, with 256 million chess positions. All training code is taken exactly from the Leela open source repository, with default hyperparameters for training. The only modification we make is encoding the player rating directly into the board state (since in the standard 112x8x8 board represntation, many of the layers are zeroed out because we don't encode board history). This allows us to train a single parameterized model that can represent human behavior at all skill levels between 400 and 2800.

\section{Exploring Human-Understandable Concepts in Our Interventions}
\label{sec:concepts}

One advantage of studying interventions in the domain of chess is that positions can be evaluated using a large number of structured, human-understandable concepts derived from classical chess evaluation functions. Modern engines such as Stockfish compute hundreds of intermediate features that capture properties of a position such as material balance, piece activity, pawn structure, and king safety. While our intervention strategy itself does not directly use these concepts, they provide a useful lens for examining the types of positions in which interventions occur.

\subsection{Chess Concepts}

In this section, we perform a descriptive analysis of intervention states using a subset of evaluation features derived from the Stockfish evaluation function. Specifically, we consider ten middlegame evaluation components which correspond to the highest-level intermediate terms used within the Stockfish evaluation function:

\squishlist
\item \textbf{PieceValue:} The total material value of pieces on the board based on predefined piece weights.
\item \textbf{Imbalance:} Adjustments to material evaluation based on the interaction of different piece types.
\item \textbf{BishopPair:} A bonus awarded when a player possesses both bishops.
\item \textbf{Pawns:} Evaluation of pawn structure features such as isolated, doubled, or supported pawns.
\item \textbf{Pieces:} Evaluation of piece placement and coordination.
\item \textbf{Mobility:} The number of legal squares available to pieces, reflecting piece activity and freedom of movement.
\item \textbf{Threats:} Tactical pressure on opposing pieces, including attacks on undefended or weak targets.
\item \textbf{PassedPawn:} Bonuses associated with pawns that have no opposing pawns blocking their path to promotion.
\item \textbf{Space:} Control of central territory and safe squares in the opponent's half of the board.
\item \textbf{King:} Evaluation of the king's defensive structure and exposure to potential attacks.
\squishend

\subsection{Methodology}

We focus on the multiple-intervention setting described in Section 4.3 and analyze positions encountered under a policy tuned to produce an intervention frequency of approximately 5\%. For each position encountered during simulated play, we compute the value of each Stockfish concept listed above.

To simplify analysis, we discretize each concept into three categories—low, medium, and high—based on its value in the position, based on the difference in evaluation scores for the two players. For example, PieceValue-High indicates that the player to move has more material than the opponent, PieceValue-Med indicates that both players have the same material, and PieceValue-Low indicates that the player to move has less material than the opponent.

We then compare the distribution of these categories between positions where the intervention policy overrides the human policy, and positions where no intervention occurs.
For each concept and rating level, we compute the change in category frequency between intervention states and non-intervention states. Intuitively, a large difference indicates that the intervention policy correlates with positions characterized by that concept.

In Figure \ref{tab:concept_interventions}, we present the top ten concepts with the largest difference between intervention and non-intervention states for each player rating.

\begin{table*}[t]
\centering
\setlength{\tabcolsep}{4pt}
\resizebox{\textwidth}{!}{

\begin{tabular}{ll|ll|ll|ll|ll}
\toprule
\multicolumn{2}{c|}{\textbf{800}} &
\multicolumn{2}{c|}{\textbf{1200}} &
\multicolumn{2}{c|}{\textbf{1600}} &
\multicolumn{2}{c|}{\textbf{2000}} &
\multicolumn{2}{c}{\textbf{2400}} \\
\textbf{Concept} & $\Delta$ &
\textbf{Concept} & $\Delta$ &
\textbf{Concept} & $\Delta$ &
\textbf{Concept} & $\Delta$ &
\textbf{Concept} & $\Delta$ \\
\midrule

King-High & 0.1862 &
PieceValue-Med & 0.0824 &
PieceValue-Med & 0.1166 &
Threats-High & 0.1006 &
PieceValue-Med & 0.1116 \\

Mobility-High & 0.0848 &
Threats-High & 0.0740 &
Threats-High & 0.0892 &
PieceValue-Med & 0.0692 &
PassedPawn-High & 0.0708 \\

PieceValue-Med & 0.0780 &
King-High & 0.0744 &
Mobility-Med & 0.0656 &
Pieces-High & 0.0548 &
Threats-High & 0.0640 \\

Threats-High & 0.0734 &
Mobility-Med & 0.0654 &
Pawns-Low & 0.0638 &
King-High & 0.0540 &
Pawns-Low & 0.0576 \\

Mobility-Med & 0.0514 &
Pawns-Low & 0.0512 &
Threats-Low & 0.0598 &
Imbalance-Low & 0.0456 &
Mobility-Med & 0.0416 \\

Pieces-High & 0.0464 &
Imbalance-Low & 0.0436 &
Imbalance-High & 0.0528 &
Mobility-Med & 0.0446 &
Threats-Low & 0.0382 \\

Imbalance-High & 0.0450 &
Threats-Low & 0.0346 &
Pieces-Low & 0.0454 &
Pawns-High & 0.0438 &
Imbalance-High & 0.0340 \\

PieceValue-High & 0.0368 &
BishopPair-High & 0.0278 &
Pawns-High & 0.0368 &
Threats-Low & 0.0396 &
Pieces-Low & 0.0336 \\

BishopPair-High & 0.0314 &
Pieces-High & 0.0216 &
Pieces-High & 0.0316 &
Imbalance-High & 0.0366 &
Pieces-High & 0.0318 \\

Space-High & 0.0284 &
Pieces-Low & 0.0214 &
King-Low & 0.0308 &
BishopPair-High & 0.0364 &
Pawns-High & 0.0282 \\

\bottomrule
\end{tabular}
}

\caption{Top concepts whose frequency differs most between intervention and non-intervention states for each rating level in the 5\% intervention frequency setting. Concepts are derived from the Stockfish evaluation function and discretized into Low, Medium, and High categories. $\Delta$ denotes the absolute difference in frequency between intervention and non-intervention positions.}

\label{tab:concept_interventions}

\end{table*}

While this analysis is purely descriptive, it suggests several preliminary patterns. For 800-rated players, the largest distribution shift is for King-High, indicating that intervention states are often positions in which the opponent king is especially weak. This is consistent with the possibility that interventions at low skill levels frequently help players find decisive continuations in already promising attacking positions. At higher ratings, however, king-safety related concepts are less dominant. One possible explanation is that stronger players are already more capable of converting positions with exposed kings without assistance, so interventions become relatively more associated with other tactical or positional features.

These results should be interpreted cautiously. Because the intervention strategy itself does not explicitly use these concepts, the analysis does not imply that the model has learned explicit chess concepts or that these features causally determine intervention decisions. Instead, the analysis provides a descriptive view of the types of positions where interventions tend to occur.

Nevertheless, such concept-based analyses may provide a useful diagnostic tool for examining intervention behavior. In particular, viewing intervention decisions through the lens of established chess concepts may help identify systematic patterns in the model’s behavior. More broadly, this type of analysis may offer a starting point for future work exploring how intervention strategies relate to interpretable domain knowledge, or how similar analyses could be used to diagnose weaknesses in human decision-making for educational or training purposes.

\end{document}